\documentclass[letterpaper]{article}
\usepackage{aaai18}
\usepackage{times}
\usepackage{helvet}
\usepackage{courier}
\usepackage{amsfonts}
\usepackage{amsmath}
\usepackage{multirow}
\usepackage{graphicx}
\begin{document}
%
\title{Image Quality Assessment Techniques Show Improved Training and Evaluation of Autoencoder Generative Adversarial Networks}
\author{Michael O. Vertolli \and Jim Davies \\
Carleton University\\
1125 Colonel By Dr\\
Ottawa, Canada K1S 5B6\\
}
\maketitle
\begin{abstract}
\begin{quote}
We propose a training and evaluation approach for autoencoder Generative Adversarial Networks (GANs), specifically the Boundary Equilibrium Generative Adversarial Network (BEGAN), based on methods from the image quality assessment literature. Our approach explores a multidimensional evaluation criterion that utilizes three distance functions: an $l_1$ score, the Gradient Magnitude Similarity Mean (GMSM) score, and a chrominance score. We show that each of the different distance functions captures a slightly different set of properties in image space and, consequently, requires its own evaluation criterion to properly assess whether the relevant property has been adequately learned. We show that models using the new distance functions are able to produce better images than the original BEGAN model in predicted ways.
\end{quote}
\end{abstract}

\section{Introduction}

\begin{figure}[htpb]
\includegraphics[scale=0.392]{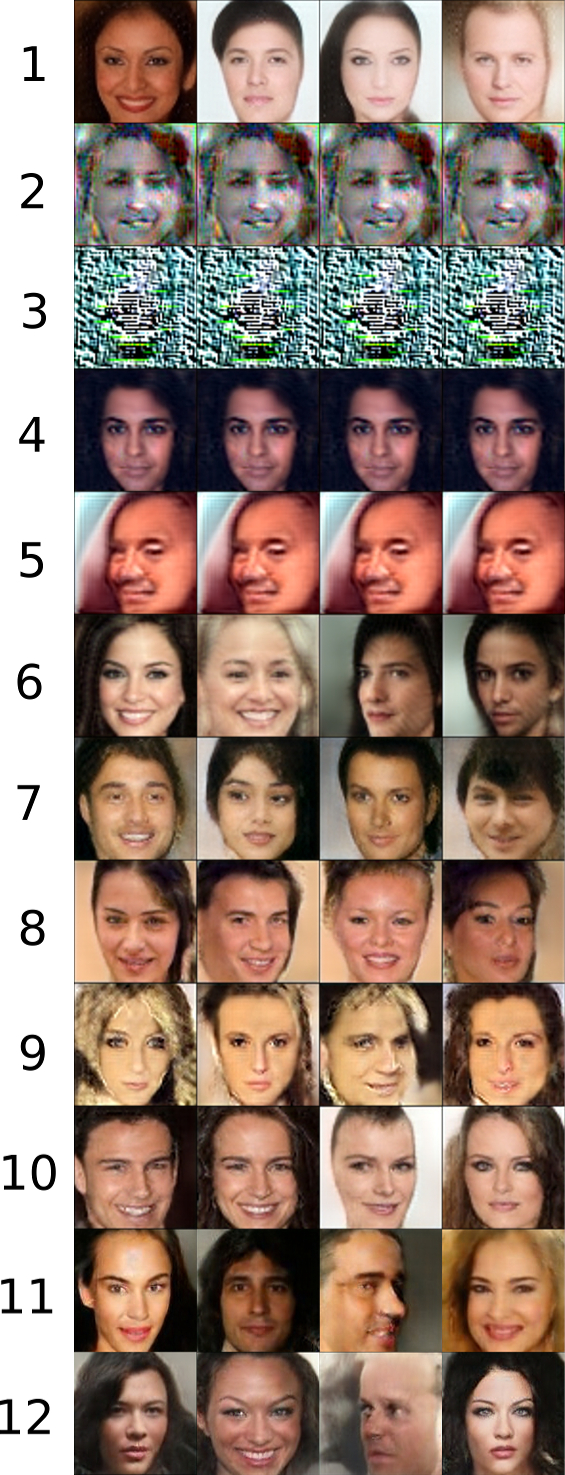}
\caption{Four outputs of each of the generators of all 12 models. The best images for each model were hand-picked. The first row is model 1, which corresponds with the original BEGAN model. Rows 2-12 represent our experiments. 
Each cell represents the output of a random sample.}
\label{fig:allimgs}
\end{figure}

\noindent Generative adversarial networks (GANs) are a class of machine learning algorithms that are designed to sample from and model a data distribution \cite{goodfellow2014gan}. In this respect, GANs function as one class of implicit probabilistic models that define a stochastic procedure that is able to directly generate data \cite{mohamed2016implicitgans}. These models are ``implicit'' because the data of interest and its associated probability density function cannot be specified explicitly.

What is unique to implicit models and their associated problems is that they are likelihood-free \cite{gutmann2016likelihoodfree,mohamed2016implicitgans}. As a consequence of the domain of interest, the relationship between the model's parameters and the data it generates is analytically intractable and, consequently, it is not possible to specify the likelihood of the model parameters given the data. This means that implicit models cannot be evaluated using standard probabilistic inference and parameter learning techniques, which rely on a likelihood, log-likelihood, Kullback-Leibler divergence, or similar functions. In short, implicit models cannot be trained using standard machine learning techniques, or, in other words, classifier training does not work well for procedural generation.

Instead, GANs leverage a form of two-sample or hypothesis testing that uses a classifier, called a \textit{discriminator}, to distinguish between observed (training) data and data generated by the model or \textit{generator}. Training is then simplified to a competing (i.e., adversarial) objective between the discriminator and generator, where the discriminator is trained to better differentiate training from generated data, and the generator is trained to better trick the discriminator into thinking its generated data is real.

GANs have been very successful in a number of domains, including but not limited to image classification/clustering \cite{radford2015dcgan}, image blending \cite{wu2017gpgan}, image in-painting \cite{li2017inpaintinggan}, super-resolution \cite{ledig2016superresolutiongan}, text classification \cite{miyato2016textclassificationgan}, and text-to-image synthesis \cite{zhang2016stackgan}. The bulk of this research is focused on the generation of 2D images or image parts. Recent developments (see Figure \ref{fig:allimgs}, row 1) on a sub-family of GANs, called autoencoder GANs (AE-GANs), are able to produce near photo-realistic results \cite{berthelot2017began}.

Problematically, the challenges of using implicit models resurface when evaluating the output produced by these models. In the image domain, \citeauthor{theis2015geneval} (\citeyear{theis2015geneval}) have shown that the assumption of model-data convergence is not guaranteed for these models, resulting in unpredictable effects. In particular, they demonstrate that log-likelihood scores are decoupled from the quality of the images generated by the model. Largely as a consequence of this issue, the GAN literature has taken to evaluating the quality of the resulting images through visual inspection images or interpolations across the latent space of the model (see Figure \ref{fig:interps}).

In what follows, we expand on the existing GAN literature by using image quality assessment techniques. We show how these techniques are constructed and how they can be used to improve the evaluation of AE-GAN model output in the image domain. We also show how the components of these techniques can be used to improve training for AE-GANs, specifically the boundary equilibrium GAN (BEGAN) model. First, we will give a more detailed specification of the AE-GAN architecture.

\section{Related Work}

\subsection{Autoencoder GANs}

In the original GAN specification, the task is to learn the generator's distribution $p_{G}$ over data $\boldsymbol{x}$ \cite{goodfellow2014gan}. To accomplish this, one defines a generator function $G(\boldsymbol{z}; \theta_{G})$, which produces an image using a noise vector $\boldsymbol{z}$ as input, and $G$ is a differentiable function with parameters $\theta_{G}$. The discriminator is then specified as a second function $D(\boldsymbol{x}; \theta_D)$ that outputs a scalar representing the probability that $\boldsymbol{x}$ came from the data rather than $p_G$. $D$ is then trained to maximize the probability of assigning the correct labels to the data and the image output of $G$ while $G$ is trained to minimize the probability that $D$ assigns its output to the fake class, or $1 - D(G(z))$. Although $G$ and $D$ can be any differentiable functions, we will only consider deep convolutional neural networks in what follows.

The insight of the Autoencoder GANS (AE-GANs) is that $D$ can be expanded from a single-dimensional criterion---the scalar class probability---to a multidimensional criterion by constructing it as an autoencoder \cite{zhao2016ebgan}. The image output by the autoencoder can then be compared directly to the output of $G$ using mean squared error (MSE) or other distance functions.

Recent work on AE-GANs has shifted to a comparison of autoencoder loss distributions---rather than sample distributions---using the Wasserstein distance and an equilibrium hyper-parameter \cite{berthelot2017began}. The autoencoder loss $\mathcal{L}: \mathbb{R}^{N_x} \mapsto \mathbb{R}^+$ is defined as: 

\begin{equation}
\mathcal{L} = d(v, D(v))
\label{eq:loss}
\end{equation}

\noindent where $v$ is a sample from $p_G$, and $d$ is a distance function. 
The resulting objective for the BEGAN model is:

\begin{equation}
\begin{cases}
\mathcal{L}_D = \mathcal{L}(x) - k_t \cdot \mathcal{L}(G(z)) & \text{for } \theta_D\\
\mathcal{L}_G = \mathcal{L}(G(z)) & \text{for } \theta_G\\
k_{t+1} = k_t + \lambda_k(\gamma\mathcal{L}(x) - \mathcal{L}(G(z))) & \text{for each } t \\
\end{cases}
\label{eq:began}
\end{equation}

\noindent where $k_t \in [0, 1]$ is the emphasis put on $\mathcal{L}(G(z))$ at training step $t$ for the gradient of $D$, $\lambda_k$ is the learning rate for $k$, and $\gamma \in [0, 1]$. The $\gamma$ hyper-parameter is set to relax the equilibrium between the expected value of the loss of real data and the expected value of the loss of the generator's output as follows: 

\begin{equation}
\mathbb{E}[\mathcal{L}(G(z))] = \gamma\mathbb{E}[\mathcal{L}(x)]
\end{equation}

In the early stages of training, the generator's output is easy for the autoencoder to reproduce, due to its low quality. Its loss is near 0 and $\mathcal{L}(x) > \mathcal{L}G(z))$. This equilibrium constraint guarantees this relationship is preserved over the course of training. This, in turn, guarantees that there is a valid error signal for the discriminator as the generator's output improves and, consequently, becomes harder to approximate by the autoencoder.

One major weakness of this approach comes from the selection of $d$ in Equation \ref{eq:loss}. The standard approach is to use MSE or the $l^1$ or $l^2$ norm on the difference of $v$ and $D(v)$. However, research in the image quality assessment (IQA) literature has shown that both of these metrics poorly approximate quality estimates of human raters. In the next section, we introduce a collection of techniques that can better inform how we construct $d$. This, in turn, can improve how we train and evaluate GANs overall.

\subsection{Image quality assessment}
Image quality assessment (IQA) is an area of research that focuses on evaluating the quality of digital images \cite{wang2006iqabook}. IQA is usually divided into two branches. The subjective branch evaluates images using human judgment and the objective IQA research models human judgement with computational methods. 

Objective IQA has three sub-areas: full-reference, no-reference, and reduced reference \cite{wang2006iqabook}. Full-reference IQA is  used when a `true' or `perfect' quality version (the reference of the image) to be evaluated exists. No-reference IQA has no perfect image, and reduced reference IQA only has access to certain properties or features of the reference image. This work focuses on full-reference techniques, as they have a very natural correspondence to the function $d$ from Equation \ref{eq:loss}. For example, the standard norms and MSE are all full-reference techniques. However, no-reference and reduced-reference IQA might be productively explored in future work.

\begin{figure}[htpb]
\includegraphics[scale=0.12]{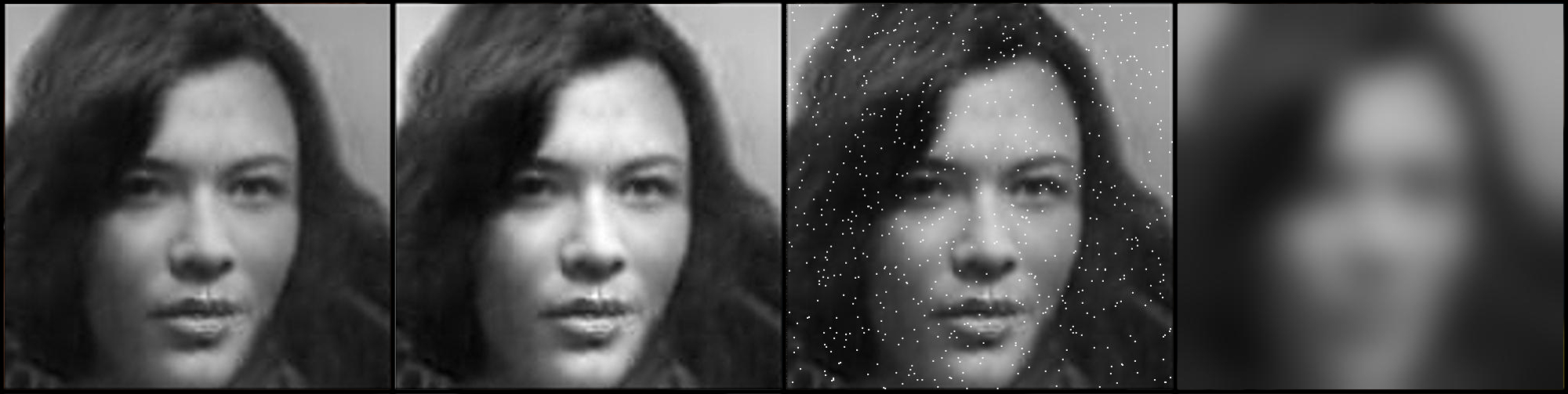}
\caption{From left to right, the images are the original image, a contrast stretched image, an image with impulsive noise contamination, and a Gaussian smoothed image. Although these images differ greatly in quality, they all have the same MSE from the original image. They are all on the mean square error (MSE) hypersphere with an MSE of approximately 400.}
\label{fig:msesphere}
\end{figure}

Bottom-up approaches try to directly model human judgement using the error visibility or error sensitivity paradigm \cite{wang2004ssim,wang2006iqabook}. This paradigm assumes that the loss of quality in an image is directly related to the visibility of the error signal. The simplest example of this paradigm is MSE, which quantifies the strength of the error signal. Problematically, if we think of the error signal as a vector from the original image, MSE actually defines a hypersphere of images in the total image space that describes images that vary greatly in visual quality (see Figure \ref{fig:msesphere}). As a consequence, it is generally accepted in the IQA literature that MSE is a poor indicator of image quality \cite{wang2006iqabook}. Bottom-up approaches suffer from a number of known problems despite a number of improvements that they have made in quantifying the error signal based on psychophysical experiments \cite{wang2004ssim,wang2006iqabook}. The most fundamental of these is the quality definition problem. It is not clear that error visibility is strongly correlated with image quality. At best, only a moderate empirical correlation has been shown \cite{silverstein1996imagequality}. For this reason, we restrict ourselves to top-down approaches, which treat human judgement as a black-box.

The top-down approach to full-reference IQA seeks to model human judgement based on a few global assumptions. For example, the Structure Similarity (SSIM) index family of models assumes that local structures are important for determining quality \cite{wang2003msssim,wang2004ssim,wang2011infweightiqa}. Many different sets of assumptions exist in this family with many different models for each of them, including MS-SSIM which is briefly mentioned below. As another example, the information-theoretic family of models assumes that the amount of information preserved between the reference (the perfect image) and the evaluated image predicts image quality \cite{sheikh2005ifc,sheikh2006vif}. A number of detailed reviews and comparisons of state-of-the-art models exist \cite{lin2011iqasurvey,zhang2012friqareview,chandler2013iqareview}. In what follows, we chose to examine an extension of the Gradient Magnitude Similarity Deviation (GMSD) model as a candidate for $d$ in  Equation \ref{eq:loss}. This model, called the color Quality Score (cQS) currently has some of the best IQA scores on a number of IQA databases and image distortion types \cite{gupta2017qs}.

\subsection{Gradient Magnitude Similarity}

Gradient Magnitude Similarity Deviation (GMSD) is a kind of structure preserving model, like SSIM \cite{xue2014gmsd}, and shares their assumptions. One way to understand these assumptions is in terms of modifications to a reference image vector ($\boldsymbol{x}$), where each dimension corresponds to a pixel value. 

One can think of modifications to the reference image as distortion vectors that are added to it. As we stated above, distortion vectors of equal length define a hypersphere of images that have an equal MSE in the image space, but can have very different perceptual qualities \cite{wang2006iqabook}. This is why the length of a distortion vector is a poor measure of quality. Other properties of these distortion vectors \textit{might} be informative, such as direction \cite{wang2006iqabook}.

SSIM characterizes images in terms of luminance, contrast, and structure. One of the good things about SSIM's notion of structure is that it does not change with variations in luminance and contrast. 
One of the key insights of this approach is that structural distortions are equivalent to rotations of a luminance-contrast plane in image space \cite{wang2006iqabook}. 

SSIM's similarity function acts on the luminance, contrast, and structure components of an image individually and is reminiscent of the Dice-Sorensen similarity function.\footnote{The Dice-Sorensen distance function does not satisfy the triangle inequality for sets \cite{gragera2016semimetric}. Since sets are a restricted case for Equation \ref{eq:sim}, where all the values are either 0 or 1, we can conclude that the corresponding distance of Equation \ref{eq:sim} also fails to satisfy the triangle inequality. Consequently, it is not a true distance \textit{metric}.}
For image component values $v_1$ and $v_2$ the function is defined as follows:

\begin{equation}
S(\boldsymbol{v}_1, \boldsymbol{v}_2) = \frac{2\boldsymbol{v}_1\boldsymbol{v}_2 + C}{\boldsymbol{v}_1^2 + \boldsymbol{v}_2^2 + C}
\label{eq:sim}
\end{equation}

\noindent where C is a constant, and all multiplications occur element-wise \cite{wang2006iqabook}.\footnote{We use $C = 0.0026$ following the work on cQS described below \cite{gupta2017qs}.} This function has a number of desirable features. It is symmetric (i.e., $S(v_1, v_2) = S(v_2, v_1)$, bounded by 1 (and 0 for $x > 0$), and it has a unique maximum of 1 only when $v_1 = v_2$. It is also consistent with Weber's law, which emphasizes the importance of change relative to a context \cite{wang2006iqabook}.

GMSD was informed by work on SSIM \cite{xue2014gmsd}. In particular, G-SSIM shifted the application of the contrast and structural similarities to the gradient of the image, rather than the image pixels \cite{chen2006gssim}. GMSD simplifies this process by simply comparing local gradients directly across two images with Equation \ref{eq:sim}. This equates to comparing the edges between images. To accomplish this, a $3\times 3$ Prewitt filter for the horizontal and vertical directions is convolved with the reference (r) and distorted (d) images and then collapsed to the gradient magnitude (m) of each image. The gradient magnitude similarity (GMS) is then computed as follows:

\begin{equation}
GMS(\boldsymbol{m}_r, \boldsymbol{m}_d) = \frac{2\boldsymbol{m}_r\boldsymbol{m}_d + C}{\boldsymbol{m}_r^2 + \boldsymbol{m}_d^2 + C}
\end{equation}

\noindent The GMS is then pooled into a scalar value using either the mean (GMSM) or standard deviation (GMSD).\footnote{Note that in the original GMSD the gradient is calculated over a small window of the total image. As our images were very small, we considered the entire image to be a single window.}

Both GMSD and SSIM were originally designed to compute similarity scores on grayscale images. \citeauthor{gupta2017qs} (\citeyear{gupta2017qs}) extended the GMS approach to apply to color images. The resulting color Quality Score (cQS) is comparable or outperforms GMSD on three standard databases, although the two models produce similar results \cite{gupta2017qs}. 

The cQS adds two additional computations to GMSM and uses a Sobel filter instead of Prewitt \cite{gupta2017qs}. To perform these computations, it first converts images to the YIQ color space model. In this model, the three channels correspond to the luminance information (Y) and the chrominance information (I and Q). GMSM is then computed on the Y channel and Equation \ref{eq:sim} is computed directly on the I and Q channels, individually. The average of the product of the modified I and Q channels is then summed with the GMSM score. 

Because of its performance, and its ability to compare color images, our method uses cQS.

\subsection{Other GAN evaluations}

There have been few attempts to improve the evaluation of GAN output. The inception score has been used by some researchers to compare and evaluate models \cite{salimans2016inceptionscore,berthelot2017began,rosca2017vaeaegan}. We chose not to include the inception score in our analyses because it is a type of no-reference image quality assessment, which is not as amenable to GAN training. Another approach used the multi-scale structural similarity score (MS-SSIM) to evaluate diversity in an image set \cite{odena2016mssimgan,rosca2017vaeaegan}. Although the comparison between images (real or generated) is an interesting application of image quality assessment research, it is very different from the comparison of loss distributions. In the current research, we compare the loss distribution of the generated images before and after autoencoding to the loss distribution of real images before and after autoencoding. Finally, the standard structural similarity score has been used to evaluate the quality of images \cite{zhao2017multiview,juefei2017ssimgan}. Our research expands on this basic idea both theoretically and in terms of application.

\section{Method}

Recall that the goal is to extend and improve both how autoencoder GANs (AE-GANs) are evaluated and how they are trained. At present, we have described the AE-GAN architecture, shown that the distance function ($d$) in Equation \ref{eq:loss} is a poor approximation of image quality, and discussed alternate methods from the full-reference IQA literature that could be good candidates for $d$. Finally, we chose the color quality score (cQS) as the primary candidate as it is state-of-the-art, simple, efficient, and designed to process color images. 

Previous image quality assessment techniques used in the GAN literature often rely on simple, single component distance measures, such as MSE, $l_1$ norm, or the $l_2$ norm. These do not correlate well with human judgement. Our hypothesis is that a weighted set of compontents will better model human judgement, and as such be a better function of image quality to be used in training and evaluation of GANs. Image quality then emerges via the interaction of a set of (ideally) orthogonal properties that adequately models human judgement.


In order to use multiple components, one needs to be able to assess multiple models that differ in how they weigh each component.
BEGANs do not compare images directly, but rather compare the loss distribution of the generated images before and after autoencoding to the loss distribution of real images before and after autoencoding. This allows one to assess the extent to which any given component has been lost in the autoencoded generated images relative to their autoencoded real counterparts. In this sense, the loss of any component in the autoencoded real images acts as a guide for how much of that component one would expect to lose if the generated image was equivalent to the real image for that component. This approach is a powerful means to deduce how much of a given component should, ideally, be present in some image.


We created several models with different component weightings. To do this we trained a series of BEGAN models with $d$ set to a weighted multidimensional function $\mathcal{L}^\mathcal{D}: \mathbb{R}^{N_x} \mapsto [0, 1]$ of a set $\mathcal{D}$ of $d$s. The resulting function is defined as follows:

\begin{equation}
\mathcal{L}^\mathcal{D} = \frac{\sum_{d\in\mathcal{D}} d(v, D(v))\beta_d}{\sum_{d\in\mathcal{D}}\beta_d}
\label{eq:multiloss}
\end{equation}

\noindent where $\beta_d$ is the weight that determines the proportion of each $d$ to include for a given model, and $\mathcal{D}$ includes the $l_1$ norm, GMSM, and the chrominance part of cQS as individual $d$s.\footnote{The `distance' of GMSM and chrominance is defined as one minus the corresponding similarity score.} We then changed the values of $\beta_d$ in order to evaluate the effects of each distance function.

A complete list of all the different models and their associated $\beta_d$ parameters is available in Table \ref{tbl:modeltypes}. They are as follows. Models 1, 7, and 11 are the original BEGAN model. Models 2 and 3 only use the GMSM and chrominance distance functions, respectively. Models 4 and 8 are the BEGAN model plus GMSM. Models 5 and 9 use all three distance functions (BEGAN+GMSM+Chrom). Models 6, 10, and 12 use a 'scaled' BEGAN model ($\beta_{l_1} = 2$) with GMSM. All models with different model numbers but the same $\beta_d$ values differ in their $\gamma$ values or the output image size.

\subsection{Model architecture}

All of the models we evaluate in this paper are based on the architecture of the BEGAN model \cite{berthelot2017began}. Both the discriminator and generator are convolutional deep neural networks. The discriminator is a typical autoencoder composed of a deep encoder and decoder. The generator mirrors the architecture of the discriminator's decoder with different weights. Convolutions are $3 \times 3$ in size with exponential linear units applied at their outputs. Each convolution layer is repeated twice. Convolution filters are increased linearly with each up or down-sampling. Down-sampling is implemented by increasing the stride of the convolution to 2 and nearest neighbour is used for up-sampling. Fully-connected layers with no non-linearities map to and from the hidden state $\boldsymbol{h} \in \mathbb{R}^{N_h}$ of the discriminator. The input state is a uniform sample of $\boldsymbol{z} \in [-1, 1]^{N_z}$.

\section{Experiments}

We conducted extensive quantitative and qualitative evaluation on the CelebA dataset of face images \cite{liu2015celeba}. This dataset has been used frequently in the past for evaluating GANs \cite{radford2015dcgan,zhao2016ebgan,chen2016infogan,liu2016cogan}.\footnote{The original BEGAN model uses an undisclosed dataset of 360K celebrity faces that is more extensive than CelebA. Consequently, our results are not identical to theirs on an equivalent model.} We evaluated 12 different models in a number of combinations (see Table \ref{tbl:modeltypes}).


\begin{table}
\begin{tabular}{|c|c|c|c|c|c|}
\hline
\multirow{2}{*}{Model \#} & \multicolumn{5}{|c|}{Model Parameters} \\
\cline{2-6}
& Size & \hspace{1em}$\gamma$\hspace{1em} & \hspace{1em}$l_1$\hspace{1em} & GMSM & Chrom \\
\hline
01 & 64 & 0.5 & 1 & 0 & 0 \\
\hline
02 & 64 & 0.5 & 0 & 1 & 0 \\
\hline
03 & 64 & 0.5 & 0 & 0 & 1 \\
\hline
04 & 64 & 0.5 & 1 & 1 & 0 \\
\hline
05 & 64 & 0.5 & 1 & 1 & 1 \\
\hline
06 & 64 & 0.5 & 2 & 1 & 0 \\
\hline
07 & 64 & 0.7 & 1 & 0 & 0 \\
\hline
08 & 64 & 0.7 & 1 & 1 & 0 \\
\hline
09 & 64 & 0.7 & 1 & 1 & 1 \\
\hline
10 & 64 & 0.7 & 2 & 1 & 0 \\
\hline
11 & 128 & 0.7 & 1 & 0 & 0 \\
\hline
12 & 128 & 0.7 & 2 & 1 & 0 \\
\hline
\end{tabular}
\caption{Models and their corresponding model distance function parameters. The $l_1$, GMSM, and Chrom parameters are their respective $\beta_d$ values from Equation \ref{eq:multiloss}.}
\label{tbl:modeltypes}
\end{table}

\subsection{Setup}

Our setup was similar to that of BEGAN \cite{berthelot2017began}. We trained the models using Adam with a batch size of 16, $\beta_1$ of 0.5, $\beta_2$ of 0.999, and an initial learning rate of 0.00008, which decayed by a factor of 2 every 100,000 epochs. 

Parameters $k_t$ and $k_0$ were set at 0.001 and 0, respectively (see Equation \ref{eq:began}). The $\gamma$ parameter was set relative to the model (see Table \ref{tbl:modeltypes}).

Most of our experiments were performed on $64 \times 64$ pixel images with a single set of tests run on $128 \times 128$ images. The number of convolution layers were 3 and 4, respectively, with a constant down-sampled size of $8\times 8$. We found that the original size of 64 for the input vector ($N_z$) and hidden state ($N_h$) resulted in modal collapse for the models using GMSM. However, we found that this was fixed by increasing the input size to 128 and 256 for the 64 and 128 pixel images, respectively. We used $N_z = 128$ for all models except 12 (scaled BEGAN+GMSM), which used 256. $N_z$ always equaled $N_h$ in all experiments. 

Models 2-3 were run for 18,000 epochs, 1 and 4-10 were run for 100,000 epochs, and 11-12 were run for 300,000 epochs. Models 2-4 suffered from modal collapse immediately and 5 (BEGAN+GMSM+Chrom) suffered from modal collapse around epoch 65,000 (see Figure \ref{fig:allimgs} rows 2-5).

\begin{table}
\begin{tabular}{|c|c|c|c|c|c|c|}
\hline
\multirow{3}{*}{Model \#} & \multicolumn{6}{|c|}{Discriminator Loss Statistics} \\
\cline{2-7}
& \multicolumn{2}{|c|}{$l_1$} & \multicolumn{2}{|c|}{GMSM} & \multicolumn{2}{|c|}{Chrom} \\
\cline{2-7}
& $M$ & $\sigma$ & $M$ & $\sigma$ & $M$ & $\sigma$ \\
\hline
01 & \textbf{0.12} & 0.02 & 0.24 & 0.02 & 0.68 & 0.08 \\
\hline
02 & 0.90 & 0.09 & \textbf{0.17} & 0.03 & 0.99 & 0.01 \\
\hline
03 & 0.51 & 0.02 & 0.52 & 0.04 & \textbf{0.46} & 0.08 \\
\hline
\hline
04 & 0.11 & 0.01 & \textbf{0.16} & 0.02 & 0.75 & 0.07 \\
\hline
05 & 0.13 & 0.02 & 0.20 & 0.02 & \textbf{0.41} & 0.05 \\
\hline
06 & \textbf{0.10} & 0.01 & 0.17 & 0.02 & 0.69 & 0.07 \\
\hline
\hline
07 & \textbf{0.10} & 0.01 & 0.22 & 0.01 & 0.63 & 0.08 \\
\hline
08 & 0.11 & 0.01 & \textbf{\textit{0.16}} & 0.02 & 0.83 & 0.07 \\
\hline
09 & 0.13 & 0.02 & 0.20 & 0.02 & \textit{\textbf{0.42}} & 0.06 \\
\hline
10 & \textbf{0.10} & 0.01 & 0.17 & 0.02 & 0.72 & 0.08 \\
\hline
\hline
11 & 0.09 & 0.01 & 0.29 & 0.01 & \textbf{0.58} & 0.08 \\
\hline
12 & \textbf{\textit{0.08}} & 0.02 & \textbf{0.17} & 0.02 & 0.63 & 0.08 \\
\hline
\end{tabular}
\caption{Lists the models, their discriminator mean error scores, and their standard deviations for the $l_1$, GMSM, and chrominance distance functions over all training epochs. Bold values show the best scores for a given distance function for similar models indicated by the double lines. Bold and italic values are the best scores for a given distance function overall, excluding models that suffered from modal collapse. Which components are important depends on the image domain the GAN is trained on. These results suggest that model training should be customized to emphasize the relevant components.}
\label{tbl:modelstats}
\end{table}

\subsection{Evaluations}

We performed four evaluations. First, to evaluate whether and to what extent the models were able to capture the relevant properties of each associated distance function, we compared the mean and standard deviation of the error scores. We calculated them for each distance function over all epochs of all models. We chose to use the mean rather than the minimum score as we were interested in how each model performs as a whole, rather than at some specific epoch. All calculations use the distance, or one minus the corresponding similarity score, for both the gradient magnitude and chrominance values. 

Second, we qualitatively examined the difference in gradient magnitude scores between models 11 (BEGAN) and 12 (scaled BEGAN+GMSM) and chrominance scores of 9 (BEGAN+GMSM+Chrom), 11, and 12 on a small set of images.

Third, we compared the convergence measure scores for models 11 and 12 across all 300,000 epochs (see Figure \ref{fig:measure}; \citeauthor{berthelot2017began} \citeyear{berthelot2017began}). The convergence measure is defined as follows

\begin{equation}
\mathcal{M}_{global} = \mathcal{L}(x) + |\gamma\mathcal{L}(x) - \mathcal{L}G(z))|
\end{equation}

\noindent where the loss is defined as per Equation \ref{eq:multiloss}. Due to the variance in this measure, we applied substantial Gaussian smoothing ($\sigma = 0.9$) to enhance the main trends. The output of a single generated image is also included for every 40,000 epochs, starting with epoch 20,000 and ending on epoch 300,000.

Finally, we perform a qualitative evaluation of the latent space of models 11 and 12, using linear interpolation in $\boldsymbol{z}$. This is a standard technique in the GAN literature to show that the model is able to generalize from the training set, rather than just memorizing it.

Reduced pixelation is an artifact of the intensive scaling for image presentation (up to $4\times$). All images in the qualitative evaluations were upscaled from their original sizes using cubic image sampling so that they can be viewed at larger sizes. Consequently, the apparent smoothness of the scaled images is not a property of the model.

\subsection{Results}

GANs are used to generate different types of images. Which image components are important depends on the domain of these images. Our results suggest that models used in any particular GAN application should be customized to emphasize the relevant components---there is not a one-size-fits-all component choice. We discuss the results of our four evaluations below.

\subsubsection{Means and standard deviations of error scores}

Results were as expected: the three different distance functions captured different properties of the underlying images. We compared all of the models in terms of their means and standard deviations of the error score of the associated distance functions (see Table \ref{tbl:modelstats}). In particular, each of models 1-3 only used one of the distance functions and had the lowest error for the associated function (e.g., model 2 was trained with GMSM and has the lowest GMSM error score). Models 4-6 expanded on the first three models by examining the distance functions in different combinations. Model 5 (BEGAN+GMSM+Chrom) had the lowest chrominance error score and Model 6 (scaled BEGAN+GMSM) had the lowest scores for $l_1$ and GMSM of any model using a $\gamma$ of 0.5.

For the models with $\gamma$ set at 0.7, models 7-9 showed similar results to the previous scores. Model 8 (BEGAN+GMSM) scored the lowest GMSM score overall and model 9 (BEGAN+GMSM+Chrom) scored the lowest chrominance score of the models that did not suffer from modal collapse. For the two models that were trained to generate $128\times 128$ pixel images, model 12 (scaled BEGAN+GMSM) had the lowest error scores for $l_1$ and GMSM, and model 11 (BEGAN) had the lowest error score for chrominance. Model 12 had the lowest $l_1$ error score, overall.

\begin{figure}[htpb]
\includegraphics[scale=0.32]{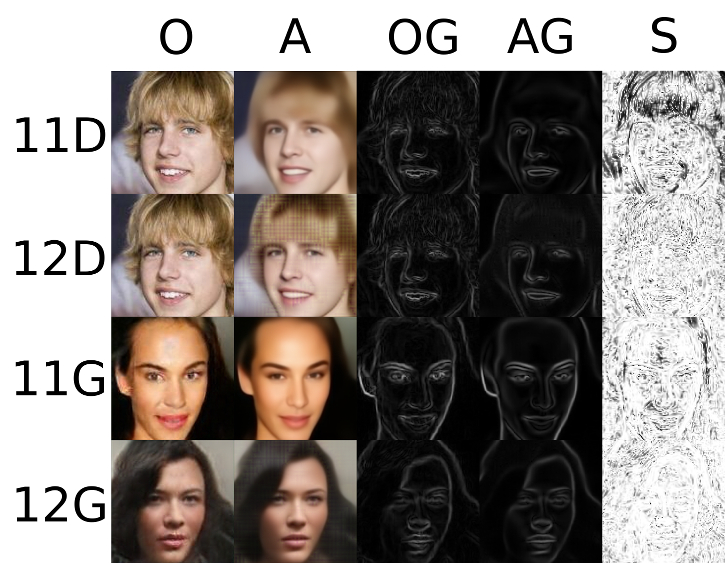}
\caption{Comparison of the gradient (edges in the image) for models 11 (BEGAN) and 12 (scaled BEGAN+GMSM), where O is the original image, A is the autoencoded image, OG is the gradient of the original image, AG is the gradient of the autoencoded image, and S is the gradient magnitude similarity score for the discriminator (D) and generator (G). White equals greater similarity (better performance) and black equals lower similarity for the final column.}
\label{fig:qualgms}
\end{figure}

\subsubsection{Visual comparison of similarity scores}

Subjective visual comparison of the gradient magnitudes in column S of Figure \ref{fig:qualgms} shows that there are more black pixels for model 11 (row 11D) when comparing real images before and after autoencoding. This indicates a lower similarity or greater loss of information in the autoencoder. Model 12 (row 12D) has a higher similarity between the original and autoencoded real images as indicated by fewer black pixels. 
This pattern continues for the generator output (rows 11G and 12G), but with greater similarity between the gradients of the original and autoencoded images than the real images (i.e., fewer black pixels overall).

\begin{figure}[htpb]
\includegraphics[scale=0.18]{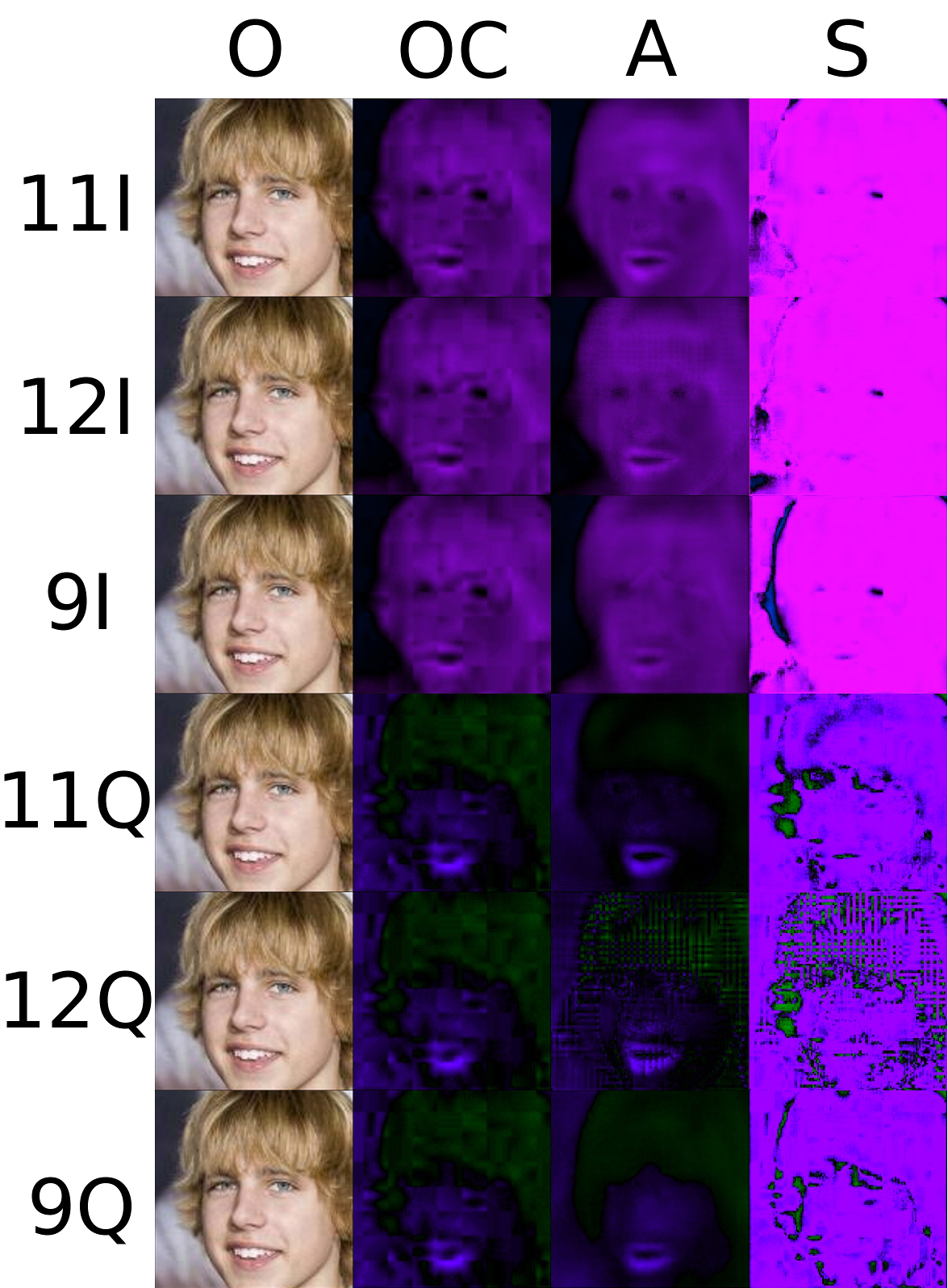}
\caption{Comparison of the chrominance for models 9 (BEGAN+GMSM+Chrom), 11 (BEGAN) and 12 (scaled BEGAN+GMSM), where O is the original image, OC is the original image in the corresponding color space, A is the autoencoded image in the color space, and S is the chrominance similarity score. I and Q indicate the (blue-red) and (green-purple) color dimensions, respectively. All images were normalized relative to their maximum value to increase luminance. Note that pink and purple approximate a similarity of 1, and green and blue approximate a similarity of 0 for I and Q dimensions, respectively. The increased gradient `speckling' of model 12Q suggests an inverse relationship between the GMSM and chrominance distance functions.}
\label{fig:qualchrom}
\end{figure}

The visual comparison of chrominance and related similarity score also weakly supported our hypotheses (see Figure \ref{fig:qualchrom}). All of the models show a strong ability to capture the I dimension (blue-red) of the YIQ color space, but only model 9 (BEGAN+GMSM+Chrom) is able to accurately capture the relevant information in the Q dimension (green-purple).

\subsubsection{Diversity of latent space}

\begin{figure}[htpb]
\includegraphics[scale=0.316]{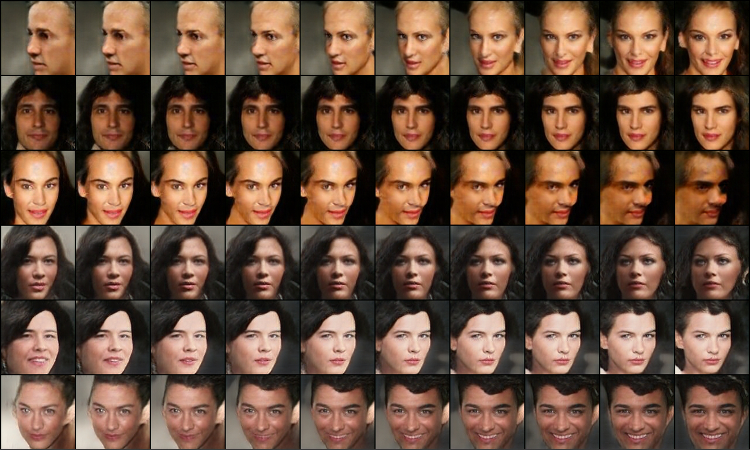}
\caption{The introduction of these new distance functions did not prevent the models from being able to produce linear interpolations in the latent space of $\boldsymbol{z}$. The top three rows are from model 11 (BEGAN) and the bottom three are from model 12 (scaled BEGAN+GMSM).}
\label{fig:interps}
\end{figure}

Further evidence that the models can generalize, and not merely memorize the input, can be seen in the linear interpolations in the latent space of $\boldsymbol{z}$. In Figure \ref{fig:interps} models 11 (BEGAN) and 12 (scaled BEGAN+GMSM) show smooth interpolation in gender, rotation, facial expression, hairstyle, and angle of the face.

\subsubsection{The BEGAN convergence measure}

\begin{figure}[ht]
\includegraphics[scale=0.071]{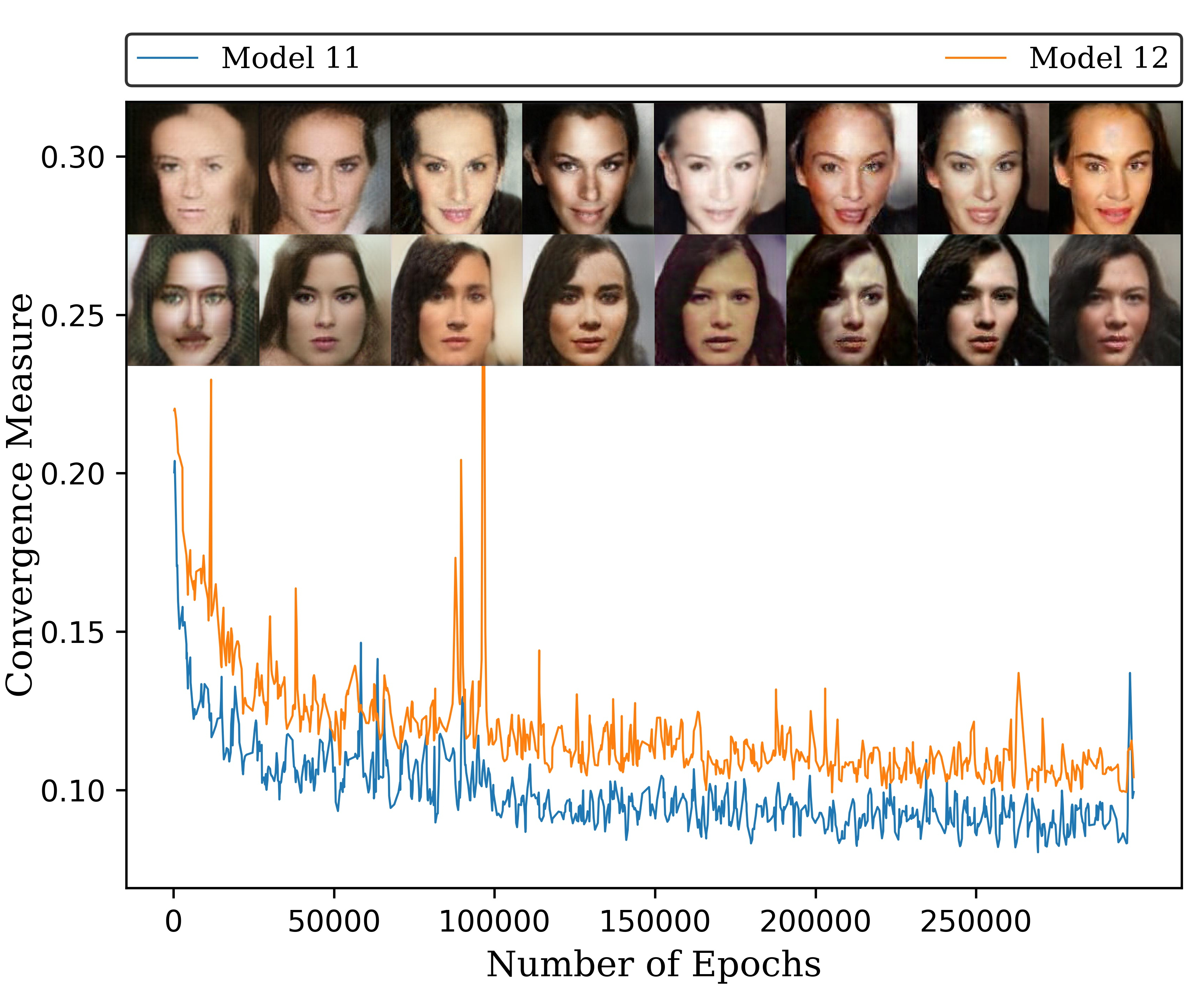}
\caption{Quality of the results for Models 11 (BEGAN) and 12 (scaled BEGAN+GMSM) with respect to the measure of convergence. The results were smoothed with a Gaussian with $\sigma = 0.9$. Images are displayed in 40,000 epoch increments starting with epoch 20,000 and going to 300,000. The top images are from Model 11 and the bottom are from Model 12. As a side note,  the output of earlier training epochs appear to be both more youthful. As training proceeds, finer details are learned by the model, resulting in apparent increased age.}
\label{fig:measure}
\end{figure}

Models 11 and 12 were compared using the convergence measure (Equation \ref{fig:measure}; \citeauthor{berthelot2017began} \citeyear{berthelot2017began}). Model 11 showed better (greater) convergence over the 300,000 epochs (as indicated by a lower convergence measure score). Both models continue to show that the convergence measure correlates with better images as the models converge.

\section{Discussion}

The goal of this project is to extend and improve both how autoencoder GANs (AE-GANs) are evaluated and trained. Our hypothesis was that the type of distance function used when computing the model's loss (Equation \ref{eq:loss}) dictates the type of properties and their associated distortions that will be acquired by the model. In a sense, these distance functions encourage the model to focus on certain properties or features about the world, much like how a parent directs their child to relevant details of a domain during learning. As a consequence of this, evaluations need to assess whether the intended features of a given distance function \textit{have been adequately acquired}.

One advantage of this approach is that it allows one to strictly define \textit{how} and \textit{in what way} one model is better than another. For example, model 12 (scaled BEGAN+GMS) is \textit{better} than model 11 (BEGAN) in that it is better able to capture the underlying gradient magnitudes of images (i.e., edges in the image). Or, to rephrase, model 12 utilizes an additional distance function whose loss it is able to approximate via learning. This view is supported by both model 12's lower GMSM error score (Table \ref{tbl:modelstats}) and the greater similarity (or smaller distortion vector) between the original and autoencoded output of both the generator and discriminator (Figure \ref{fig:qualgms}). By contrast, models 5 and 9 (BEGAN+GMSM+Chrom) use an additional chrominance distance function that they are not very good at approximating (despite having the lowest error scores for the associated property). Thus, it is not enough to include a new distance function without also evaluating whether or not a given model can learn to capture the related components. There does not seem to be any single evaluation criterion that could do this for all components in such a complex domain as that of natural images.

Interestingly, the means and standard deviations in Table \ref{tbl:modelstats} suggest that the three distance functions used are not fully independent. In particular, comparisons between models 8 (BEGAN+GMSM) and 9 (BEGAN+GMSM+Chrom) show that the introduction of the new distance function negatively impacts the two other error scores. Whether this is a consequence of the normalization value in Equation \ref{eq:multiloss} (i.e., the reduction of the proportional contribution of each distance function towards the overall score) or some other factor is not clear. The increased gradient `speckling' of model 12 (scaled BEGAN+GMSM) in color space (Figure \ref{fig:qualchrom}, row 12Q column S) also suggests an inverse relationship between the GMSM and chrominance distance functions. This is particularly strange given that the GMSM distance function only acts on the luminance dimension. One possible explanation is that the model, which is trained in the RGB color space, has difficulty fully separating the Y dimension (luminance) from the Q dimension (green-purple).

It is also very interesting that both the doubling of the contribution of the $l_1$ distance function in models 10 and 12 (scaled BEGAN+GMSM) as well as the increased $\gamma$ score in model 12 provide noticeable improvements over similar models (see Figure \ref{fig:allimgs}). We tentatively conclude that the $l_1$ distance function provides a useful boundary on the sample space of the generator, which is consistent with standard views \cite{wang2006iqabook,theis2015geneval}. This boundary can then be used to restrict the search space of other distance functions for parameter tuning, etc.

Another tentative finding is that the generator has greater similarity between the gradient magnitudes of its output before and after autoencoding than the discriminator (see Figure \ref{fig:qualgms}). In a sense, the difference between the real image before and after autoencoding is the limit (or maximum) amount of information that can be captured by that distance function. Although this might be unsurprising, it lends additional support to the loss-based approach to GAN training \cite{berthelot2017began}. That is, training GANs to better match loss distributions does have equivalent visual analogs in the space of images.

Modal collapse is a known problem with GANs \cite{berthelot2017began}. One of the standard techniques for dealing with modal collapse is to reduce the learning rate. Our experiments found that this was not particularly helpful, especially when the learning rate is already very small. Instead, we found that increasing the number of dimensions in $\boldsymbol{z}$ predictably prevented modal collapse. 

\begin{figure}[htpb]
\includegraphics[scale=0.6]{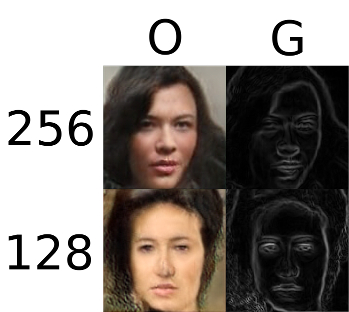}
\caption{A sample of the original output (O) and its gradient (G) for model 12 with $N_z = 256$ and $N_z = 128$. The output of 128 has a much more detail rich gradient but suffers from modal collapse. By increasing the size of $N_z$ to 256, model 12 (scaled BEGAN+GMSM) is better able to capture the relevant gradient information without suffering from modal collapse and without changing the learning rate.}
\label{fig:modcollapse}
\end{figure}

A previous instantiation of model 12 (scaled BEGAN+GMSM), whose input size ($N_z$) was only 128 dimensions, started generating minor variations on a detail-rich face just before modal collapse (see Figure \ref{fig:modcollapse}, row 128). What seems to have happened is the model restricted its domain in order to more accurately model the properties of that domain. In other words, the loss distribution of the sample was struggling to fully capture all of the complexity of the loss distribution of the data with the space that was available to it. By increasing $N_z$, we expand the sample loss distribution such that it is better able to capture the relevant complexity without suffering from modal collapse. We tentatively conclude that our addition of edge detection through the GMSM distance function \textit{with a sufficiently large input dimension size} ($N_z$) allows model 12 (scaled BEGAN+GMSM) to learn more of the fine details of the image's structure without modal collapse.

This further supports our hypothesis that the inclusion of different distance metrics changes the underlying data distribution. After all, the original BEGAN model only used 64 dimensions ($N_z = 64$) while producing high quality output that did not suffer from modal collapse. Thus, the addition of the GMSM distance function appears to have not only changed but enriched the underlying data loss distribution in well-defined ways.

\section{Conclusions}

We have provided preliminary evidence that different distance functions are able to select different components in the space of images. These functions can be used to both train Autoencoder Generative Adversarial Network (AE-GAN) models to better learn their associated properties, as well as evaluate whether or not those properties have in fact been learned.

We have shown evidence that the integration of full-reference Image Quality Analysis (IQA) techniques with the AE-GAN literature, especially BEGAN related models, can result in performance improvements. The recent uptake of MS-SSIM and SSIM as viable evaluation techniques is indicative of this trend. However, we do not believe that this is all there is left to do.

IQA approaches have been oriented towards selecting a single scalar value that corresponds well with subjective quality ratings. Our research suggests that multidimensional approaches are better suited to AE-GAN evaluation and training. Thus, it is important to determine which elements of various IQA techniques are amenable to this multidimensional approach rather than continuing the search for a single scalar value. For example, the chrominance similarity and associated distance functions are one area that our research indicates is in need of further study as it is a property that appears to be both difficult to learn and evaluate. To the extent that one is able to find a collection of suitable distance functions, one will be able to better approach the generation of a truly `good' image.

\bibliographystyle{aaai}
\bibliography{main}

\end{document}